\documentclass[twocolumn,conference]{IEEEtran}
\usepackage[latin9]{inputenc}
\usepackage{amsmath}
\usepackage{amssymb}
\usepackage{graphicx}
\usepackage[unicode=true,
 bookmarks=true,bookmarksnumbered=true,bookmarksopen=true,bookmarksopenlevel=1,
 breaklinks=false,pdfborder={0 0 0},pdfborderstyle={},backref=false,colorlinks=false]
 {hyperref}
\hypersetup{pdftitle={Your Title},
 pdfauthor={Your Name},
 pdfpagelayout=OneColumn, pdfnewwindow=true, pdfstartview=XYZ, plainpages=false}

\makeatletter
\usepackage[caption=false,font=footnotesize]{subfig}

\makeatother

\begin{document}

\title{Efficient Reduced-Order Models for Soft Actuators}

\author{Yue~Chen$^{\dagger}$, Kevin~C.~Galloway$^{\ddagger}$, and Isuru~S.~Godage$^{*}$}
\maketitle
\begin{abstract}
Soft robotics have gained increased attention from the robotic community
due to their unique features such as compliance and human safety.
Impressive amount of soft robotic prototypes have shown their superior
performance over their rigid counter parts in healthcare, rehabilitation,
and search and rescue applications. However, soft robots are yet to
capitalize on their potential outside laboratories and this could
be attributed to lack of advanced sensing capabilities and real-time
dynamic models. In this pilot study, we explore the use of high-accuracy,
high-bandwidth deformation sensing via fiber optic strain sensing
(FOSS) in soft bending actuators (SBA). Based on the high density
sensor feedback, we introduce a reduced order kinematic model. Together
with cubic spline interpolation, this model is able to reconstruct
the continuous deformation of SBAs. The kinematic model is extended
to derive an efficient real-time equation of motion and validated
against the experimental data. 
\end{abstract}

\section{Introduction\label{sec:Introduction}}

\begin{table}[b]
$\dagger$ Dept. of Mechanical Engineering, University of Arkansas,
Fayetteville, AR 72701. email: \href{mailto:yc039@uark.edu}{yc039@uark.edu}.
$\ddagger$ Dept. of Mechanical Engineering, Vanderbilt University,
Nashville, TN 37212. $*$ School of Computing, DePaul University,
Chicago, IL 60604. \vspace{1mm} \\ Yue~Chen and Isuru~S.~Godage
contributed equally to the work.\vspace{1mm} \\ This work is supported
in part by the National Science Foundation grant IIS-1718755.
\end{table}
The rise of bioinspired soft robotics demands highly compliant and
inherently safe actuators instead of precise and fast ones that are
desirable in traditional rigid-bodied robotics. In this paper, we
focus on soft bending actuators (SBA) that bends actively or passively,
during operation \cite{polygerinos2015modeling}. Lately, soft robots,
in the sense that \textquotedblleft completely\textquotedblright{}
deforms, have been an active area of research. Soft robots have often
been constructed from soft materials such as elastomeric polymers.
This enables continuous deformation with a few active degrees of freedom
(DoF) to form complex shapes (compare to fixed \textquotedblleft geometric
shapes\textquotedblright{} of rigid-bodied robots). Due to the passive
deformation these robots undergo, they are considered as infinite
DoF systems. But in reality, they only have a handful of actuated
DoFs, and therefore are highly underactuated. Many soft robot prototypes
have been proposed over the years which employed actuation methods
such as pneumatic, hydraulics, shape memory alloys, electroactive
polymers, and magnetic fluids. By systematically organizing SBA, these
prototypes demonstrated manipulation , snake locomotion \cite{luo2014design},
legged locomotion \cite{tolley2014resilient}, and peristaltic locomotion
\cite{lin2011goqbot}.

\begin{figure}[t]
\begin{centering}
\includegraphics[width=1\columnwidth]{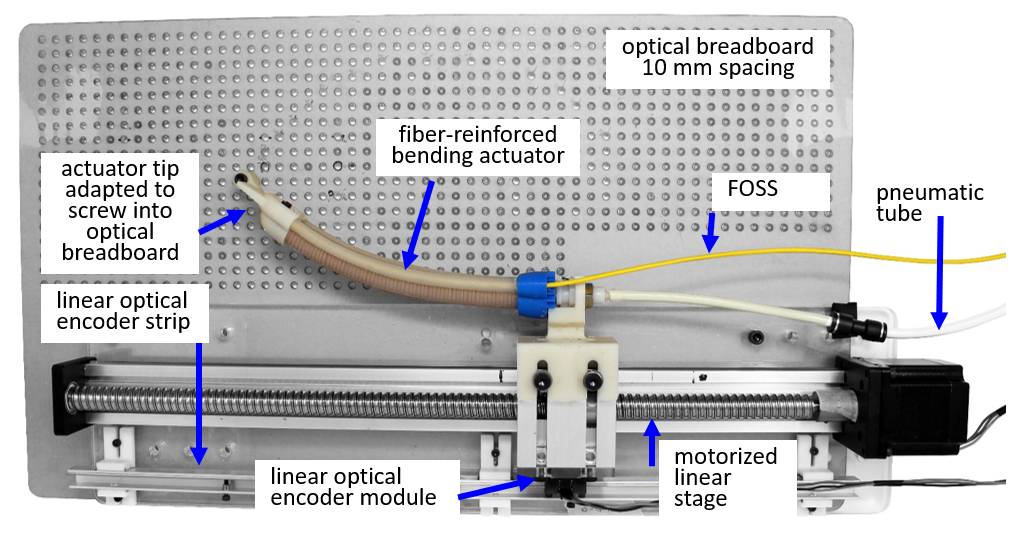}
\par\end{centering}
\caption{The prototype soft bending actuator (SBA) experimental setup with
the integrated fiber optic strain sensor (FOSS).}
\label{fig:testbed}
\end{figure}

Despite the increased amount of research in SBA, they are yet to make
an impact outside laboratory settings. The authors believe that this
is due in part to the lack of adequate models that can be used in
real-time and advanced sensing techniques (suitable for soft robots)
to provide accurate shape deformation data in real-time. Much of the
modeling approaches that have been presented to-date involves systematic
derivation of system quasi-statics based on the mechanical properties
of SBAs \cite{polygerinos2015modeling,renda2014dynamic}. These models
define the deformation as a function of pressure and thus only captures
the steady state information. Yet, pneumatic actuators are highly
compliant and therefore can deform into complex shapes as a result
of momentum of fast actuation or external forces. Thence, such static
models are of limited use in applications where speed dominates the
system behavior and raises the need for better and dynamic models. 

On top of that, the sensors utilized in soft robots so far have not
been able to reap the benefits of soft robots. For instance, image
based sensing, capable of measuring the entire shape deformation in
real-time, requires specific backgrounds and line-of-sight to be reliable
in automated image processing and thus limits the robot's application
space. Other sensing methods include highly nonlinear bending sensors
and strain sensors that are unreliable in practice. When obtaining
the state feedback, the most common approach is to measure the tip
coordinates. But, it is not a good indicator of the soft robot as
it can be in any one of state (due to compliance) out of infinitely
many possibilities. In this respect, fiber optic strain sensors (FOSS)
could serve as high-fidelity position sensors for soft robots. 

In this pilot study, a high resolution FOSS sensor array is integrated
into a fiber-reinforced SBA to obtain accurate position at 0.8~mm
along the SBA at 75~Hz. This data density is beyond what soft roboticists
have been able to collect from soft robots so far and thus opens up
new research avenues on how this shape data may be used to enable
applications previously deemed infeasible for soft robots. In addition
to introducing the FOSS for soft robotics, we also introduce a reduced-order
dynamic model, that matches the FOSS data, for SBA as a compromise
among numerical efficiency, accuracy, and complexity.

\section{Prototype Soft Bending Actuator with Integrated Sensing\label{sec:Experimental-Setup}}

This pilot study was conducted in the Luna Innovations Inc, headquarters
in Blacksburg, VA. The FOSS sensor we used in our experiment is based
on Optical Frequency Domain Reflectometry (OFDR) \cite{soller2005high}.
The readers are referred to \cite{kreger2016optical} for an in-depth
discussion of the FOSS sensing technology. Several features worth
noting about the FOSS platform are: (a) fiber optic cable is compliant,
(b) bendable in curvature as low as 10~mm, (c) can function under
significant morphological variation, (d) can detect bending and twisting
along the entire length, (e) produce full (translation and rotation)
spatial data at every 0.8~mm, (f) sampling rates up to 250~Hz, and
(g) eliminates the need for line-of-sight and thus, can be integrated
into soft material robotic structures.

\begin{figure}[t]
\begin{centering}
\includegraphics[width=1\columnwidth]{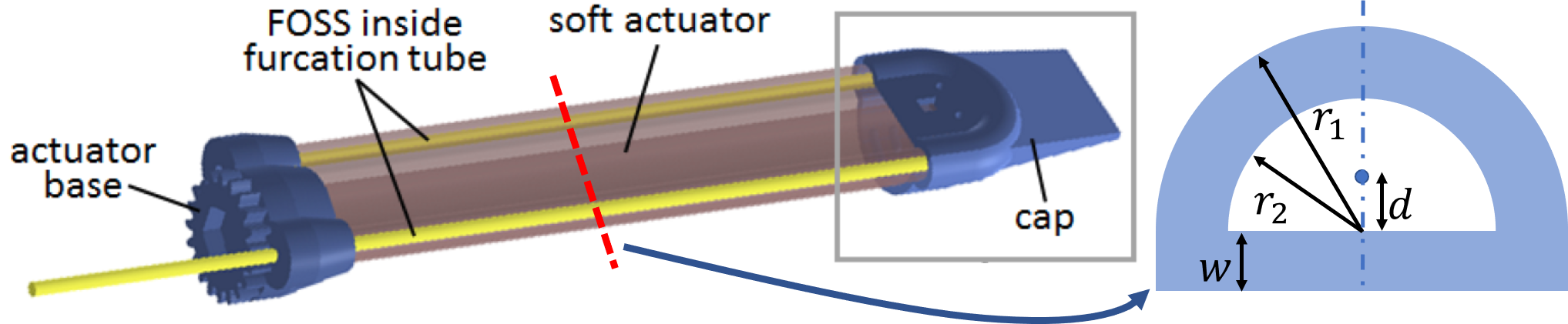}
\par\end{centering}
\caption{(Left) Integration of fiber optic strain sensor into the soft bending
actuator, and (Right) the cross-section of the SBA.}
\label{fig:softActMaking}
\end{figure}

The hardware experimental setup used in this pilot study is shown
in in Fig. \ref{fig:testbed} which is controlled via a Matlab Simulink
Realtime System. A digital proportional valve was used to supply pressure
the actuator at various pressures. A pressure sensor was used to record
the pressure readings. The details of the fabrication process of the
SBA used here (see Fig. \eqref{fig:softActMaking}) can be found in
\cite{polygerinos2015soft}. SBA bending portion is 17~cm long, weights
69~g, and has semi-annular cross section. The fiber-reinforced wall
thickness, $w$, is estimated to be $4\,mm$ with $r_{1}=14\,mm$
and $r_{2}=10\,mm$. The reinforcement fibers around the bladder increase
the operating pressure range and thus the resulting output force.
To maintain the FOSS without significantly straining and affecting
the natural compliance of SBA, we co-molded a low-friction Teflon
lined lumens around the edge of the SBA body (see Fig. \ref{fig:softActMaking}).
The FOSS was housed within a 3~mm diameter furcation tube (Fiber
Instrument Sales, Inc., Oriskany, NY, part \#: F00FR3NUY) and restrains
the sensor from bending beyond 10~mm radius of curvature. As FOSS
has no markers, it is a challenge to determine the end points of the
SBA. To solve this problem, we attached a cap with a constant curvature
at the tip of the SBA. By identifying the curvature and its length,
we could then find the tip location from the experimental data. We
then rigidly attached the SBA with the FOSS sensor to a fixed table
in the FOSS's XY plane and gather dynamic data.

\section{Methodology\label{sec:System-Model}}

\subsection{Reduced-Order Model}

\begin{figure}[tb]
\begin{centering}
\includegraphics[bb=0bp 0bp 390bp 200bp,width=0.8\columnwidth]{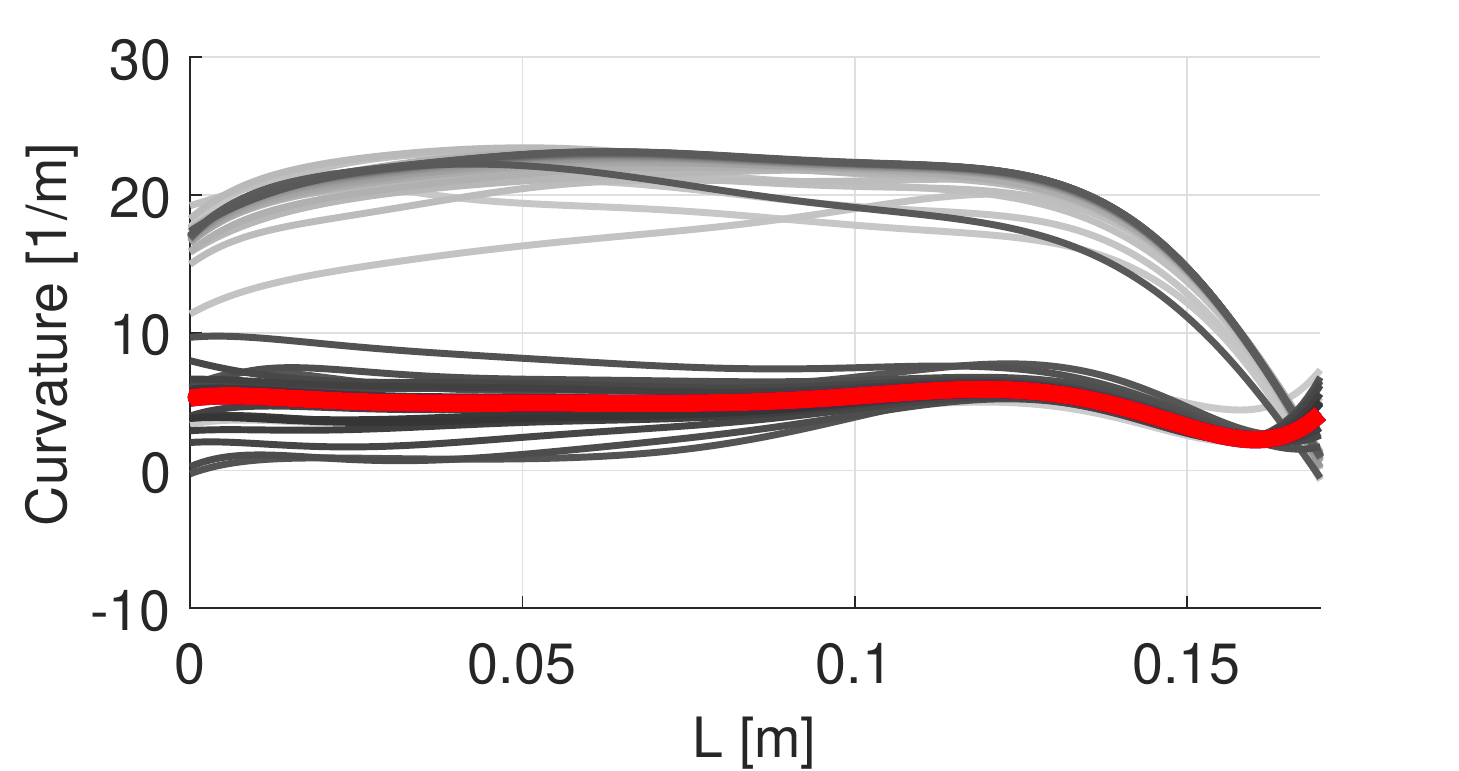}
\par\end{centering}
\caption{Curvature profile of SBA during a 240~kPa step input. }
\label{fig:curvatureVariation}
\end{figure}

To-date, SBA models consider uniform material distribution and therefore
simulate bending as circular arcs (constant curvature). However, SBA
fabrication processes are prone to variations in physical and mechanical
parameters (material density and thickness) and results in non-uniform
bending. Figure \ref{fig:curvatureVariation} shows the curvature
variation (initial value in red) along the SBA during motion. Hence,
a new model that could account for such deviations need to be proposed. 

Ideally, a high-DoF, discrete jointed system would be capable of capturing
the nonlinear deformation. But, analogous to finite element models,
they are numerically inefficient for real-time applications. Here,
we combine a low-order discrete system and cubic spline interpolation
to represent the smooth and continuous bending. We utilized experimental
data, collected from the SBA for a step response, to determine the
lowest order discrete model capable of acceptable shape reconstruction
as follows. The FOSS data at any time contains Cartesian (XYZ) trajectories
for 212 points (0.8~mm apart) along the entire length. We divided
the length of SBA to 2-5 similar-length segments (from each time instance)
and applied cubic spline interpolation to reconstruct the deformation
shape. We used the maximum Euclidean distance between the reconstructed
curve and the actual shape as the measure of similarity between the
curves. Figure \ref{fig:fitting} shows the error progression for
discrete jointed systems of 2-5 rigid segments. The 5-link system
reconstructed the bending shape with less than 3~mm maximum error
(and 1.3~mm mean error) whereas low order discretizations lead to
large errors (spikes around 4.1~s). 
\begin{center}
\begin{figure}[tb]
\begin{centering}
\includegraphics[width=0.75\columnwidth]{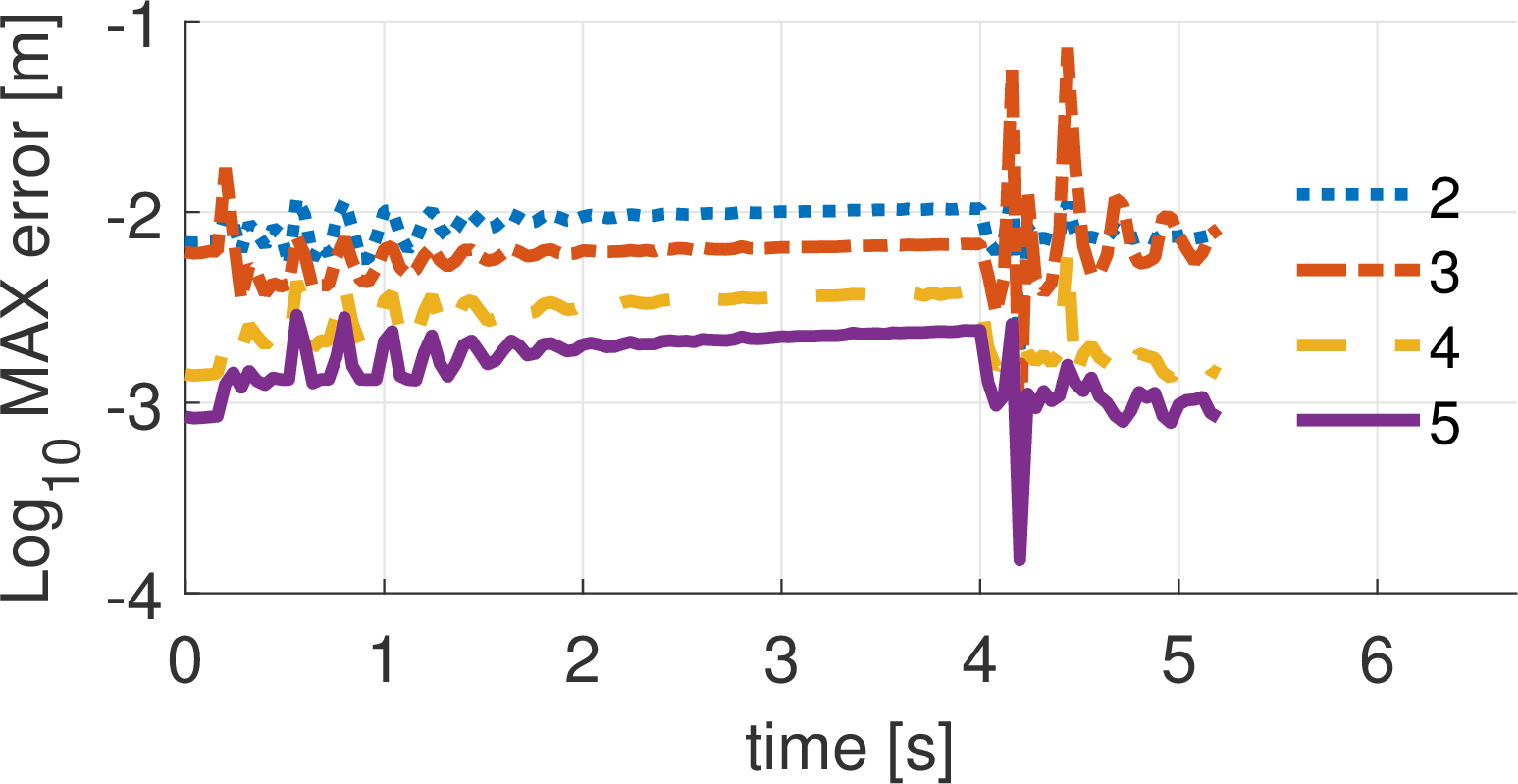}
\par\end{centering}
\caption{Maximum error comparison for 2-5 point approximation. The 5-link system
resulted 0.0024~m maximum error and 0.0012~m mean error for the
duration of the experiment.}
\label{fig:fitting}
\end{figure}
\par\end{center}

Figure \ref{fig:matchCurve} shows some instances of this experiment
where we compare the actual data (+ marks) and the reconstructed curve
(solid line) along with the discrete points we used to reconstructed
the shape. It could be seen that the shape reconstruction is identical
without any noticeable departure from the measured curve. Thus, we
will use a 5-link discrete approximation for the development of the
dynamic model.
\begin{center}
\begin{figure}[tb]
\begin{centering}
\includegraphics[bb=0bp 0bp 390bp 285bp,width=0.75\columnwidth]{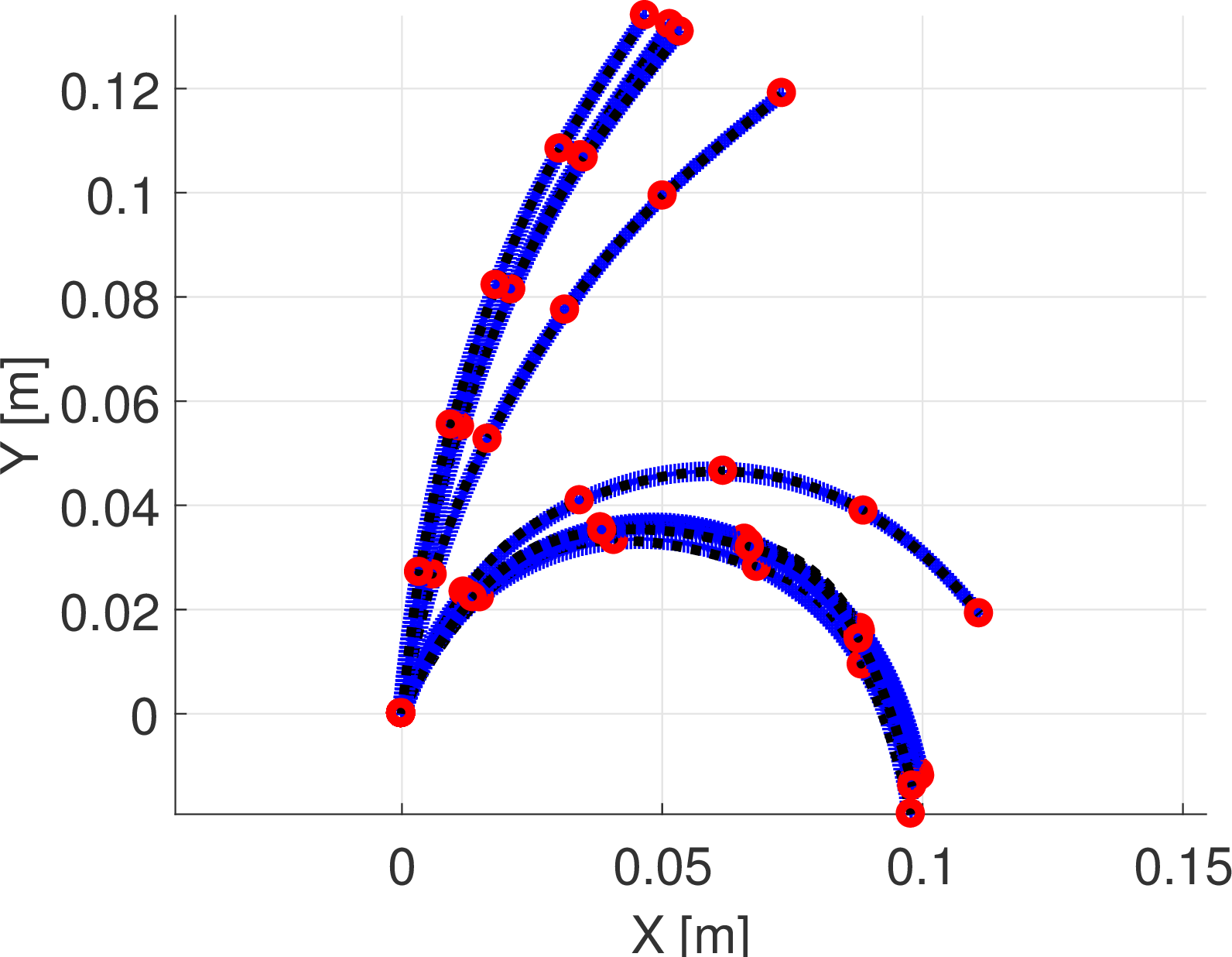}
\par\end{centering}
\caption{The difference between the measured and the reconstructed deformation
shapes of the SBA deformation during bending (Fig. \ref{fig:fitting}).
+ experimental data, o joints of the reduced-order model and the solid
line shows the cubic spline interpolation.}
\label{fig:matchCurve}
\end{figure}
\par\end{center}

\subsection{Kinematic Model\label{subsec:Kinematic-Model}}

\begin{figure}[b]
\begin{centering}
\begin{minipage}[t]{0.9\columnwidth}%
\begin{center}
\includegraphics[width=1\columnwidth]{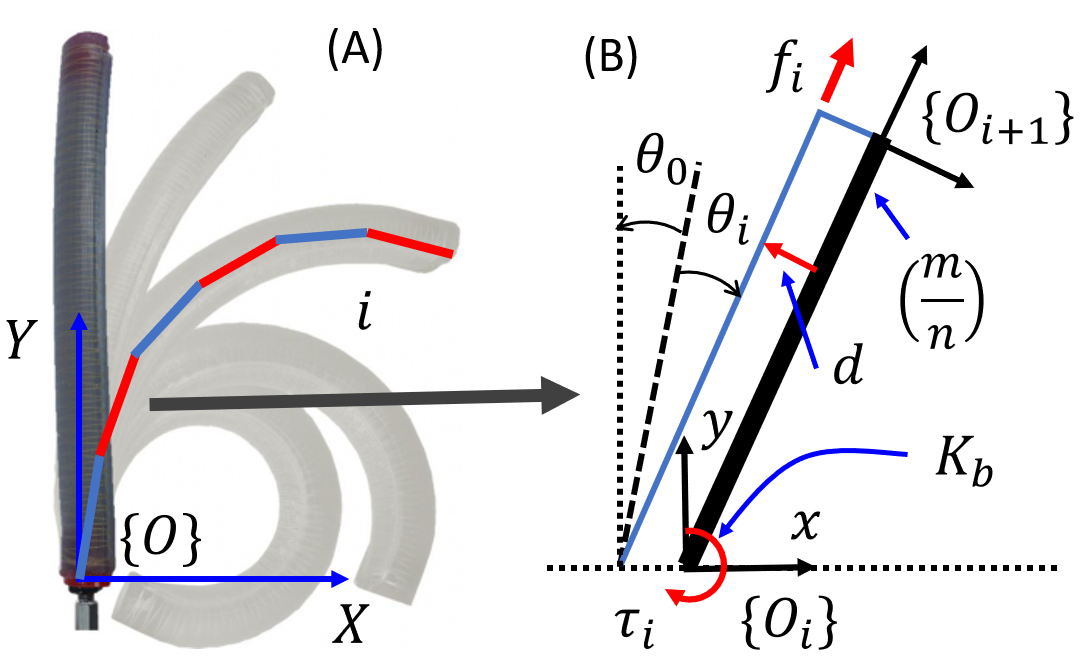}
\par\end{center}%
\end{minipage}
\par\end{centering}
\caption{(Left) the discrete jointed model of SBA in the task-space coordinate
system, $\left\{ O\right\} $, (Right) the segment parameters in the
local coordinate frame, $\left\{ O_{i}\right\} $, and related physical
properties (further detailed in Section \ref{subsec:Dynamic-Model}).}
\label{fig:schematic}
\end{figure}

The schematic of a planar 5-link discrete system model is shown in
Fig. \ref{fig:schematic}-A. Let us consider any $i^{th}$ link of
the discrete model (Fig. \ref{fig:schematic}-B). Note that, since
we define the discrete joins based on the FOSS data, the link lengths
($l_{i}$) are not identical. As shown in Fig. \ref{fig:testbed}
the unactuated SBA has a noticeable curvature (see Fig. \ref{fig:testbed}).
Thus, each link has an angle offset $\theta_{i0}$ (also not uniform)
relative to the previous segment while $\theta_{i}$ is the joint-space
variable. The proposed approach attempts to find a middle ground (with
5~DoF) to emulate an infinite DoF SBA system with minimal error while
facilitating real-time dynamics.

The standard homogeneous transformation techniques can be applied
to derive the homogeneous transformation matrix (HTM) of any $i^{th}$
segment, $\mathbf{T}_{i}\in\mathbb{SE}^{2}$ is given by \eqref{eq:Ti}
where $\mathbf{R}_{z}\in\mathbb{SO}^{2}$ is the rotational HTM about
the +Z axis and $\mathbf{P}_{y}\in\mathbb{R}$ is the translational
HTM along the +Y axis. $\mathbf{R}_{i}\in\mathbb{R}^{2\times2}$ and
$\boldsymbol{p}_{i}\in\mathbb{R}^{2\times1}$ are the rotational and
translation matrices of the segment with respect to the local coordinate
system, $\left\{ O_{i}\right\} $.

\begin{align}
\mathbf{T}_{i}\left(\theta_{i}\right) & =\mathbf{\mathbf{R}_{y}}\left(\theta_{i}+\theta_{i0}\right)\mathbf{P}_{z}\left(l_{i}\right)=\left[\begin{array}{cc}
\mathbf{R}_{i} & \boldsymbol{p}_{i}\\
0 & 1
\end{array}\right]\label{eq:Ti}
\end{align}
Employing the standard serial link kinematics, the HTM of the $i^{th}$
link, $\mathbf{T}^{i}:\boldsymbol{q}_{i}\mapsto\mathbb{SE}^{2}$,
with respect to the task-space,$\left\{ O\right\} $, is given by
\eqref{eq:T^i} where $\boldsymbol{q}_{i}=\left[\theta_{1}\cdots\theta_{i}\right]^{T}\in\mathbb{R}^{i}$
is the joint-space vector and $\mathbf{R}^{i}\in\mathbb{R}^{2\times2}$
and $\boldsymbol{p}^{i}\in\mathbb{R}^{2\times1}$ are the segment
rotational and translation matrices relative to $\left\{ O\right\} $.

\begin{align}
\mathbf{T}^{i} & =\prod_{k=1}^{n}\mathbf{T}_{k}=\left[\begin{array}{cc}
\mathbf{R}^{i} & \boldsymbol{p}^{i}\\
0 & 1
\end{array}\right]\label{eq:T^i}
\end{align}

Expanding \eqref{eq:T^i}, the recursive relationships for $\mathbf{R}^{i}$
and $\boldsymbol{p}^{i}$ can be derived as

\begin{align}
\begin{split}\mathbf{R}^{i} & =\mathbf{R}^{i-1}\mathbf{R}_{i}\\
\boldsymbol{p}^{i} & =\boldsymbol{p}^{i-1}+\mathbf{R}^{i-1}\boldsymbol{p}_{i}
\end{split}
\label{eq:Rp^i}
\end{align}
where $\mathbf{R}^{i-1}$ and $\boldsymbol{p}^{i-1}$, according to
our definition, is the tip coordinates of the prior, $\left(i-1\right)^{th}$
segment.

We take thee time derivative of \eqref{eq:Rp^i} to calculate the
body velocity (i.e., with respect to $\left\{ O_{i}\right\} $) of
the $i^{th}$ segment in recursive form as

\begin{align}
\begin{split}\boldsymbol{\Omega}_{i} & =\mathbf{R}_{i}^{T}\left(\boldsymbol{\Omega}_{i-1}\mathbf{R}_{i}+\dot{\mathbf{R}}_{i}\right)\\
\boldsymbol{\upsilon}_{i} & =\mathbf{R}_{i}^{T}\left(\boldsymbol{\upsilon}_{i-1}+\boldsymbol{\Omega}_{i-1}\boldsymbol{p}_{i}+\dot{\boldsymbol{p}}_{i}\right)
\end{split}
\label{eq:wvi_recursive}
\end{align}
where $\boldsymbol{\Omega}_{i}\in\mathbb{R}^{2\times2}$ and $\boldsymbol{\upsilon}_{i}\in\mathbb{R}^{2\times1}$
are the skew symmetric angular velocity matrix and linear velocity
vector respectively \cite{murray2017mathematical}. The angular velocity,
$\boldsymbol{\omega}\in\mathbb{R}$ can be derived from from $\boldsymbol{\Omega}_{i}$,
as $\boldsymbol{\omega}_{i}=\left[\boldsymbol{\Omega}_{i}\right]_{12}$.

\subsection{Dynamic Model\label{subsec:Dynamic-Model}}
\begin{flushleft}
The kinematic relationships derived in \ref{subsec:Kinematic-Model}
are used to derive the equations of motion (EoM) in standard form
given by \eqref{eq:eom}. A detailed treatment of the recursive numerically
efficient dynamic formulations, the readers are referred to \cite{godage2016dynamics}. 
\par\end{flushleft}

\begin{align}
\mathbf{M}\ddot{\boldsymbol{q}}+\left(\mathbf{C}+\mathbf{D}\right)\dot{\boldsymbol{q}}+K_{b}\boldsymbol{q} & =\boldsymbol{\tau}\label{eq:eom}
\end{align}
where $\mathbf{M}$, $\mathbf{C}$, $\mathbf{D}$, $K_{b}$, and $\boldsymbol{\tau}$
are the generalized inertia matrix, centrifugal/Coriolis force matrix,
damping force matrix, torsional spring coefficient at the joints,
and the joint-space torque applied to the system. From the semi-annular
cross-section of SBA bladder, we can compute the external torque,
$\tau_{i}=pAd$ where $d=\frac{4r_{2}}{3\pi}$ ($d$ shown in Fig.
\ref{fig:schematic}-B), $p$ is pressure, and $A=\frac{\pi r_{2}^{2}}{2}$
is the SBA cross-section area. $K_{b}$ and $\mathbf{D}$ are difficult
to estimate and therefore found via experimental characterization
process, similar to approach reported in \cite{godage2016dynamics}
to obtain the optimal $K_{b}=1.6067$ and $\mathbf{D}=10^{-3}\mathbf{\text{diag}}\left([8,8,8,8,8]\right)$.
The EoM is then implemented in Matlab Simulink platform and solved
using the ODE15s solver. 

\section{Experimental Results\label{sec:Experimental-Results}}

\begin{figure}[tb]
\begin{centering}
\includegraphics[width=1\columnwidth,height=0.75\columnwidth]{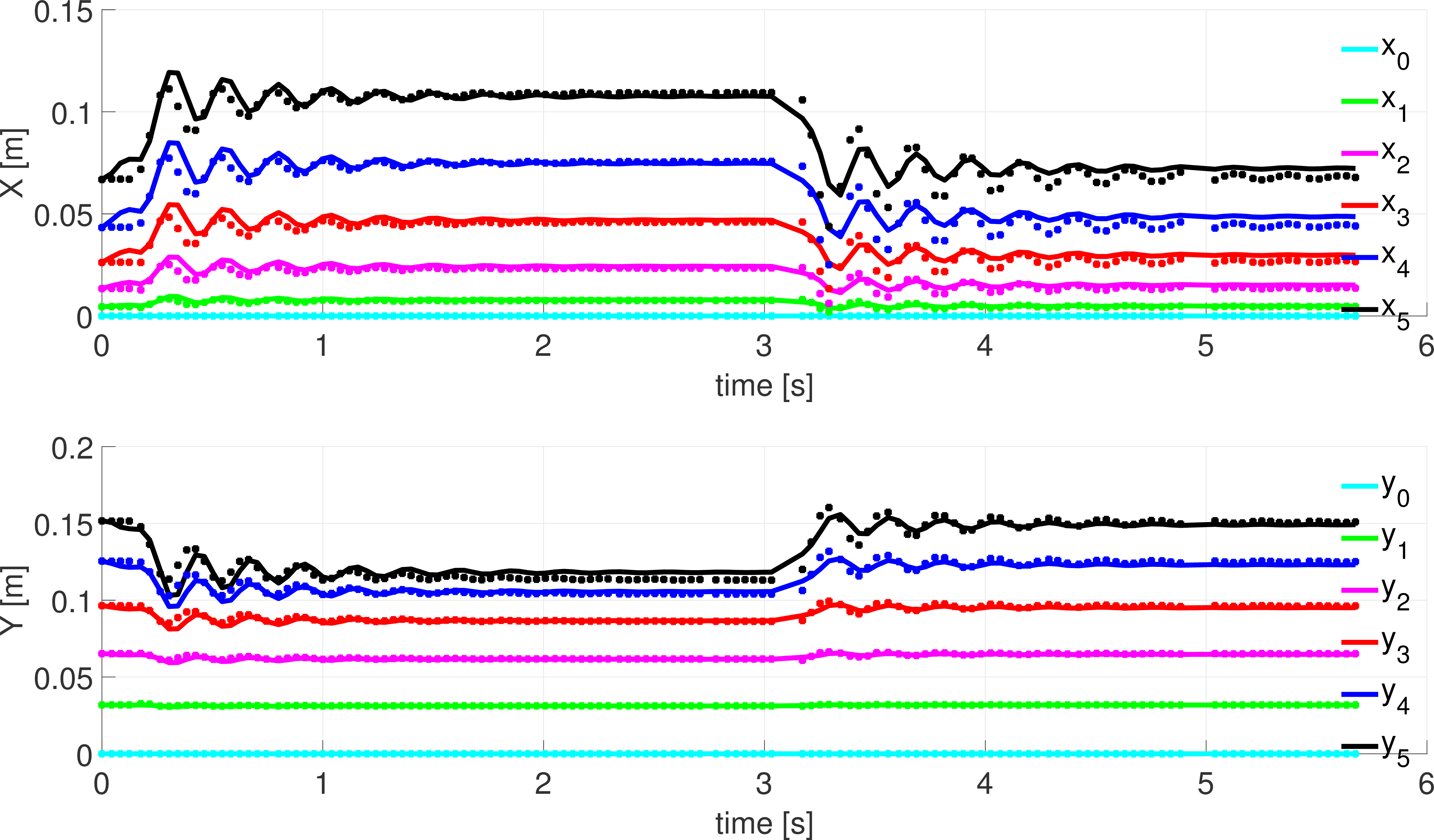}
\par\end{centering}
\caption{Comparison of the numerical results with the experimental data. The
subscripts of the X,Y coordinates denotes the task-space trajectories
of the tips of segments in the task-space,$\left\{ O\right\} $.}
\label{fig:exp_10}
\end{figure}

Figure \ref{fig:exp_10} shows the step response comparison of EoM
given in \ref{subsec:Dynamic-Model} and experimental data for a rectangular
pressure input of 119~kPa for 2.76~s duration starting at 0.12~s.
The plot shows the X and Y coordinate trajectories of the 5 tip positions
of the reduced order discrete link dynamic model (numbered from 1-5).
Empirical results demonstrates that the numerical model simulates
the SBA well overall. The abrupt pressure increase causes the SBA
to move suddenly causing the entire arm to bend and oscillate (due
to momentum and high compliance). The system is under-damped with
oscillation frequency about 5~Hz and achieve steady state due to
the friction of the system. It is worth nothing that, the SBA models
proposed so far yet to model dynamic behavior at such high frequencies.
The proposed model successfully simulates the SBA response well overall,
specifically, the overshoot and vibration frequency. In addition,
the model is able to well model the under-damped oscillations, not
only at the tip, but also along the length of the arm. We can observe
some steady stat errors towards the end of the simulation. These can
be attributed to the inherent hysteresis of SBAs. The ongoing work
explores the feasibility of incorporating the hysteretic behavior
of SBA using the model reported in \cite{godage2012pneumatic}.

\section{Conclusions and Future Work\label{sec:Conclusions}}

Soft bending actuators have shown strong potential powering the next
generation soft robotics for for applications in non-engineered and
human-friendly spaces. So far, that potential has not been realized.
This is due to the lack of sophisticated feedback systems and advanced
dynamic models. In this paper, we integrate a SBA with FOSS sensor
to obtain high fidelity data of the entire SBA up to 250~Hz rate
and demonstrate orders of magnitude better look at the SBA state-space
than the prevailing sensing techniques. To meet the demands of accuracy
and numerical efficiency, we systematically derived and proposed a
reduced-order dynamic model. The simulated output was then examined
against the FOSS data to validate the model. The proposed model exhibited
good agreement in capturing the dynamics of SBA. Future work will
focus on implementing hysteresis and dynamic control of SBAs for applications
in environmental sensing \cite{Galloway2018} and contact detection.

\bibliographystyle{IEEEtran}
\bibliography{Biblio}

% Generated by IEEEtran.bst, version: 1.14 (2015/08/26)
\begin{thebibliography}{10}
\providecommand{\url}[1]{#1}
\csname url@samestyle\endcsname
\providecommand{\newblock}{\relax}
\providecommand{\bibinfo}[2]{#2}
\providecommand{\BIBentrySTDinterwordspacing}{\spaceskip=0pt\relax}
\providecommand{\BIBentryALTinterwordstretchfactor}{4}
\providecommand{\BIBentryALTinterwordspacing}{\spaceskip=\fontdimen2\font plus
\BIBentryALTinterwordstretchfactor\fontdimen3\font minus
  \fontdimen4\font\relax}
\providecommand{\BIBforeignlanguage}[2]{{%
\expandafter\ifx\csname l@#1\endcsname\relax
\typeout{** WARNING: IEEEtran.bst: No hyphenation pattern has been}%
\typeout{** loaded for the language `#1'. Using the pattern for}%
\typeout{** the default language instead.}%
\else
\language=\csname l@#1\endcsname
\fi
#2}}
\providecommand{\BIBdecl}{\relax}
\BIBdecl

\bibitem{polygerinos2015modeling}
P.~Polygerinos, Z.~Wang, J.~T. Overvelde, K.~C. Galloway, R.~J. Wood,
  K.~Bertoldi, and C.~J. Walsh, ``Modeling of soft fiber-reinforced bending
  actuators,'' \emph{IEEE Transactions on Robotics}, vol.~31, no.~3, pp.
  778--789, 2015.

\bibitem{luo2014design}
M.~Luo, W.~Tao, F.~Chen, T.~K. Khuu, S.~Ozel, and C.~D. Onal, ``Design
  improvements and dynamic characterization on fluidic elastomer actuators for
  a soft robotic snake,'' in \emph{Technologies for Practical Robot
  Applications (TePRA), 2014 IEEE International Conference on}.\hskip 1em plus
  0.5em minus 0.4em\relax IEEE, 2014, pp. 1--6.

\bibitem{tolley2014resilient}
M.~T. Tolley, R.~F. Shepherd, B.~Mosadegh, K.~C. Galloway, M.~Wehner,
  M.~Karpelson, R.~J. Wood, and G.~M. Whitesides, ``A resilient, untethered
  soft robot,'' \emph{Soft robotics}, vol.~1, no.~3, pp. 213--223, 2014.

\bibitem{lin2011goqbot}
H.-T. Lin, G.~G. Leisk, and B.~Trimmer, ``Goqbot: a caterpillar-inspired
  soft-bodied rolling robot,'' \emph{Bioinspiration \& biomimetics}, vol.~6,
  no.~2, p. 026007, 2011.

\bibitem{renda2014dynamic}
F.~Renda, M.~Giorelli, M.~Calisti, M.~Cianchetti, and C.~Laschi, ``Dynamic
  model of a multibending soft robot arm driven by cables,'' \emph{IEEE
  Transactions on Robotics}, vol.~30, no.~5, pp. 1109--1122, 2014.

\bibitem{soller2005high}
B.~J. Soller, D.~K. Gifford, M.~S. Wolfe, and M.~E. Froggatt, ``High resolution
  optical frequency domain reflectometry for characterization of components and
  assemblies,'' \emph{Optics Express}, vol.~13, no.~2, pp. 666--674, 2005.

\bibitem{kreger2016optical}
S.~T. Kreger, N.~A.~A. Rahim, N.~Garg, S.~M. Klute, D.~R. Metrey, N.~Beaty,
  J.~W. Jeans, and R.~Gamber, ``Optical frequency domain reflectometry:
  principles and applications in fiber optic sensing,'' in \emph{Fiber Optic
  Sensors and Applications XIII}, vol. 9852.\hskip 1em plus 0.5em minus
  0.4em\relax International Society for Optics and Photonics, 2016, p. 98520T.

\bibitem{polygerinos2015soft}
P.~Polygerinos, Z.~Wang, K.~C. Galloway, R.~J. Wood, and C.~J. Walsh, ``Soft
  robotic glove for combined assistance and at-home rehabilitation,''
  \emph{Robotics and Autonomous Systems}, vol.~73, pp. 135--143, 2015.

\bibitem{murray2017mathematical}
R.~M. Murray, \emph{A mathematical introduction to robotic manipulation}.\hskip
  1em plus 0.5em minus 0.4em\relax CRC press, 2017.

\bibitem{godage2016dynamics}
I.~S. Godage, G.~A. Medrano-Cerda, D.~T. Branson, E.~Guglielmino, and D.~G.
  Caldwell, ``Dynamics for variable length multisection continuum arms,''
  \emph{The International Journal of Robotics Research}, vol.~35, no.~6, pp.
  695--722, 2016.

\bibitem{godage2012pneumatic}
I.~S. Godage, D.~T. Branson, E.~Guglielmino, and D.~G. Caldwell, ``Pneumatic
  muscle actuated continuum arms: Modelling and experimental assessment,'' in
  \emph{Robotics and Automation (ICRA), 2012 IEEE International Conference
  on}.\hskip 1em plus 0.5em minus 0.4em\relax IEEE, 2012, pp. 4980--4985.

\bibitem{Galloway2018}
K.~Galloway, Y.~Chen, E.~Templeton, B.~Rife, I.~Godage, and E.~Barth, ``Fiber
  optic shape sensing with a soft actuator for state estimation and planar
  environmental mapping,'' \emph{Soft Robotics}, 2018 (under review).

\end{thebibliography}

\end{document}